\documentclass[letterpaper]{article}
\usepackage{uai2019}

\usepackage[
backend=biber,
style=authoryear,
maxcitenames=1,
maxbibnames=50,
citestyle=authoryear
]{biblatex}

\usepackage{hyperref}

\usepackage[margin=1in]{geometry}

\usepackage{caption}

\usepackage{afterpage}

\usepackage{fancyhdr,graphicx,amsmath,amssymb}
\usepackage[ruled,vlined]{algorithm2e}
\include{pythonlisting}
\SetKwInput{kwGiven}{Given}
\SetKwInput{kwwith}{With}
\usepackage{float}
\DeclareMathOperator{\E}{\mathbb{E}}
\usepackage{algorithmic}
\usepackage{hyperref}

\usepackage{times}
\usepackage[utf8]{inputenc}

\title{Curiosity-Driven Multi-Criteria Hindsight Experience Replay}

\usepackage[hang,flushmargin]{footmisc}

\author{ {\bf John B. Lanier\thanks{\hspace{4mm}Department of Computer Science\newline \hspace{4mm}University of California, Irvine}} \\
\href{mailto:jblanier@uci.edu}{jblanier@uci.edu} \\
\And
{\bf Stephen McAleer\footnotemark[2]}  \\
\href{mailto:smcaleer@uci.edu}{smcaleer@uci.edu} \\
\And
{\bf Pierre Baldi\footnotemark[2]}   \\
\href{mailto:pfbaldi@ics.uci.edu}{pfbaldi@ics.uci.edu} \\
}

\begin{document}

\maketitle

\begin{abstract}
Dealing with sparse rewards is a longstanding challenge in reinforcement learning. The recent use of hindsight methods have achieved success on a variety of sparse-reward tasks, but they fail on complex tasks such as stacking multiple blocks with a robot arm in simulation. Curiosity-driven exploration using the prediction error of a learned dynamics model as an intrinsic reward has been shown to be effective for exploring a number of sparse-reward environments. We present a method that combines hindsight with curiosity-driven exploration and curriculum learning in order to solve the challenging sparse-reward block stacking task. We are the first to stack more than two blocks using only sparse reward without human demonstrations.
\end{abstract}

\section{Introduction}

Goal-based reinforcement learning has become an important framework for formulating and solving goal-based sequential decision making tasks. 
%While a typical reinforcement learning agent uses a policy conditioned on observations to maximize cumulative reward, a goal-based agent chooses its actions based on both observations and a given goal. 
In goal-based reinforcement learning, the agent’s rewards are usually dependent on achieving a goal, and it chooses its actions using a goal-conditioned policy. Goal-conditioned policies can enable a reinforcement learning agent to generalize to new goals after training on a many different goals in the same environment (\cite{rauber2017hindsight}). 
%This is particularly useful in real-world robotics applications in which specific goals can change frequently within the same environment. 
%In such cases, the desired end result is usually clearly defined, but the steps to get there may not be (such as moving to a certain position or relocating an object). 

Goal-based reinforcement learning environments can be given a binary and sparse reward that is encountered only when the goal is reached. Defining reward in this way ensures that if the agent maximizes reward then it also reaches the user's intended goal, which is not necessarily true of manually-shaped dense rewards (\cite{leveraging}). However, sparse rewards are also difficult to learn from. As the length of a sparse-reward task increases, it becomes less likely that an agent will discover how to reach its goal through random exploration (\cite{riedmiller2018learning}). This problem is exacerbated when a sparse reward depends on the fulfillment of multiple goals or criteria. 
%, This can be done automatically across many different goals and situations, and the alternative of manually shaping dense rewards can be difficult and lead to unintended consequences. 

\begin{figure}%
    \centering
    \includegraphics[width=\columnwidth]{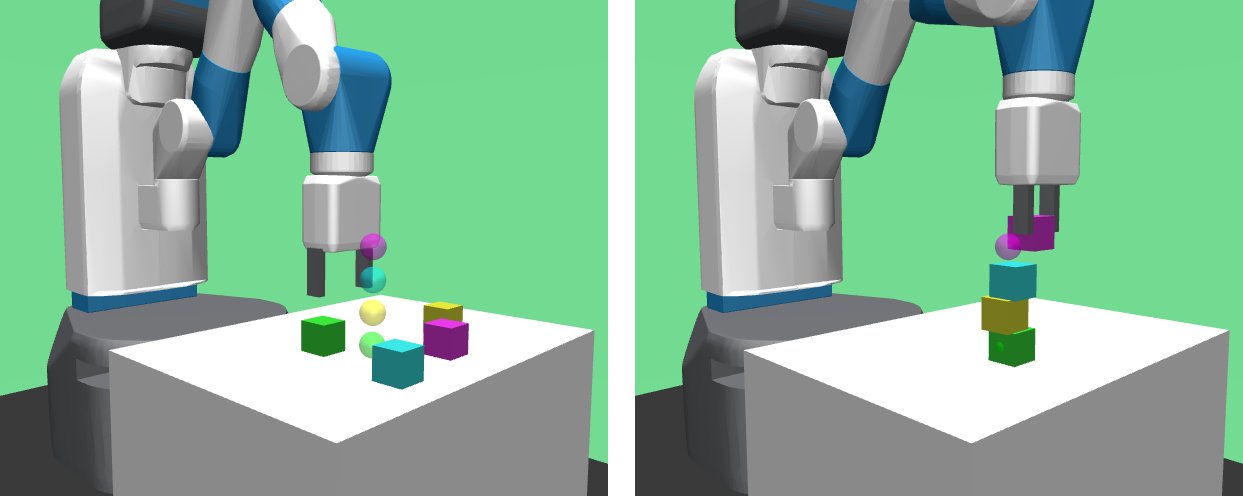}
    \caption{The simulated robotic block stacking environment. A goal consists of target positions where the blocks need to be placed, shown here as colored spheres. In the incremental reward environment, the agent receives a reward for each block being in its target position. In the sparse reward environment, the agent receives a reward only when every block is in its target position.}%
    \label{fig:example}%
\end{figure}

Recently, hindsight methods have served as a popular solution to sparse-reward goal-oriented learning by training an agent on the goals that it actually reached in addition to those which were intended (\cite{andrychowicz2017hindsight}). This is done in the hope that knowledge of how to reach randomly discovered goals will allow an agent to generalize well enough to find its assigned goals. However, in many environments, an agent can be asked to reach goals that are very different from those it may discover by chance, causing such generalization to be difficult. In these cases, the same sparse reward issues remain, making it challenging for an agent to learn how to accomplish its given objectives.

Stacking multiple blocks in a simulated robotics environment is a sparse-reward, goal-based task that highlights shortcomings of hindsight learning. Multiple-block stacking is too difficult for established hindsight methods like Deep Deterministic Policy Gradients with Hindsight Experience Replay (DDPG+HER) to reliably solve without access to human demonstrations (\cite{nair2018overcoming}). Satisfying all criteria of a block stacking goal requires learning multiple skills to correctly place each block, and the end goals are very different from those that the agent may discover with random exploration. Achieving reward by correctly placing all blocks is precarious and requires long chains of specific actions. Therefore, under a sparse reward, even with hindsight, it is highly unlikely that an agent will discover the complex sequences of actions required to place every block in its correct position on the stack. Our method is the first method that is able to solve sparse-reward block stacking for more than two blocks without access to human demonstrations. 

To solve sparse-reward multi-block stacking without help from demonstration, we use DDPG+HER combined with curiosity-driven exploration and curriculum learning. In order to balance improved exploration with exploitation during training, we introduce a new method of combining data from both curiosity-based and standard policies in an off-policy fashion. 
%with we train three policies towards the objectives of exploration, exploitation, and the combination of both, allowing the user to specify when and how the agent should directs its behavior. 
Additionally, we introduce a form of hindsight experience replay that is more sample efficient for multi-criteria goal-based environments. We show that the advantages introduced by each of these methods complement the others, and that the combination of all of them is necessary to solve the hardest stacking tasks. 

\subsection{Related Work}

% \subsubsection*{Sparse Rewards}

% There are a large number of ways in which the general difficulties of sparse RL have been approached. 

% One group of methods can be categorized by the exploitation of relevant prior knowledge. This may be done through curriculum learning (\cite{bengio2009curriculum, haber2018learning}) or multi-task learning (\cite{hausman2018learning,cabi2017intentional,rusu2016progressive}) by leveraging skills learned from others tasks to solve the target one. Demonstrations of successful trajectories can be used teach reinforcement learners to solve sparse environments from examples (\cite{paine2018one, nair2018overcoming, salimans2018learning, duan2017one}), and in the wide field of meta-learning, experience from previous environments may be used to create more efficient learning procedures apt for solving environments with less information (\cite{gupta2018unsupervised, wang2016learning, duan2016rl, finn2017model}).

% A second group of methods can be categorized through the use of more effective exploration to increase the chances of discovering sparse reward signals. Parameter-space noise (\cite{plappert2017parameter, plappert2017parametera}) can improve unstructured exploration by adding noise directly to a policy network, resulting in better performing stochastic behaviors. However, in very hard sparse environments, more structured and planned exploration is necessary. This is typically done by making the agent pursue some form of exploration related objective or reward. We refer to this set of methods as curiosity-driven exploration. 

\subsubsection{Curiosity-Driven Exploration}

We refer to curiosity-driven exploration as any method that attempts to drive an agent to explore trajectories which it has not visited frequently before, usually by making the agent pursue some form of exploration related objective or reward.

Curiosity-driven exploration has been approached by training agents to maximize information gain {(\cite{little2013learning, houthooft2016vime})}, pursue less visited areas using state pseudo-counts (\cite{bellemare2016unifying, ostrovski2017count}), and maximize state empowerment (\cite{gregor2016variational, mohamed2015variational, klyubin2005empowerment}).
 
 We focus on exploration by performing actions that both challenge and improve an agent's ability to model the world (\cite{kaushik2018multi, gordon2012hierarchical, schmidhuber1991possibility, schmidhuber2010formal}). We approach this by training a dynamics model on the state transitions that our agent visits and encouraging the agent to maximize the model's per-sample error on those transitions. Assuming a dynamics model is more accurate on transitions that it has seen frequently before, such an agent seeking to challenge the dynamics model should be inclined to visit new, rarely before seen state transitions. Choosing actions to directly challenge an online trained dynamics model has been shown to result in complex emergent behaviors (\cite{haber2018learning}). Using a dynamics model's error as an RL exploration reward can motivate an agent to seek out novel states, sometimes solving an environment's objective without extrinsic rewards, and combining environmental rewards with a bonus exploration reward has the potential to increase an agent's learning speed and end-performance (\cite{pathak2017curiosity, burda2018large}). On the same note, training a model to predict the output of a random function from state features and choosing actions to maximize its error helped achieve state-of-the art performance on the Montezuma's Revenge Atari domain (\cite{burda2018exploration}). 
 
 %Our method of curiosity-driven exploration draws from these works, as we train a forward-dynamics model and use its error as a reward for our agent to optionally pursue as an RL objective.

\subsubsection{Curriculum Learning in Goal-Based Tasks}

Previous applications of curriculum learning (\cite{bengio2009curriculum}) to goal based environments include training on a variety of tolerances for considering goals achieved (\cite{fournier2018accuracy}), masking certain goal dimensions to allow all such values on an axis to be sufficient for success (\cite{eppe2018curriculum}), and generating curricula that walk backwards from a predefined success state (\cite{florensa2017reverse}, \cite{mcaleer}). 

Intrinsically motivated goal exploration processes (IMGEPs) have also been used to automatically generate goals which maximize learning progress across one (\cite{forestier2017intrinsically, pere2018unsupervised, laversanne2018curiosity}) or multiple (\cite{colas2018curious}) tasks.

%We do not take an automated curriculum learning approach to solve block stacking. Instead, we train our learner on two easier hand-designed, foundational tasks before training on the target block stacking task.

\subsubsection{Hindsight methods}

Our work builds on Hindsight Experience Replay (HER) (\cite{andrychowicz2017hindsight}) as a way to effectively augment goal oriented transition samples for a replay buffer. Hindsight has also been adapted to policy gradient settings (\cite{rauber2017hindsight}).

Efforts have been made to increase the efficiency of HER by prioritizing the sampling of more relevant transitions. This has been done by attributing higher importance to transitions and trajectories in which more physical work is done by the agent (\cite{zhao2018energy}), rare goal states are achieved (\cite{zhao2018curiosity}), or higher temporal difference error is measured (\cite{deshpande2018improvements}). 

% In principal, such priority-based augmentations would not conflict with our presented multi-criteria goal sampling method, and could be combined with our work.

\subsubsection{Block Stacking}

Stacking multiple blocks with sparse rewards has been solved before using expert demonstration in (\cite{duan2017one}) and (\cite{nair2018overcoming}). Our work is a direct followup to the latter, as we solve a similar set of environments without demonstration.

Stacking only 2-blocks with sparse rewards has been solved without demonstration by training on an automatic curriculum which selects tasks from a small collection, prioritizing tasks with higher changes in learning progress (\cite{colas2018curious}) and by collecting data from multiple policies following auxiliary objectives to accomplish predefined interesting actions (\cite{riedmiller2018learning}). Dense reward robotic block stacking tasks have been solved before using both a model-based approach, PILCO, (\cite{deisenroth2011learning, deisenroth2011pilco}) and by initializing the environment at intermediate stages of the task (\cite{popov2017data}).

% Block stacking has also been studied from a logical planning standpoint in (\cite{dvzeroski2001relational, srivastava2014combined, kavraki1994probabilistic, kaelbling2012integrated}), however the environments considered in these are largely entirely different problems.

\subsection{Background}

\subsubsection{Reinforcement Learning}
We consider the standard reinforcement learning formalism in which an agent interacts with an environment $E$. The environment is fully observable, and consists of a set of states $S$, a set of actions $A$, a reward function $r: S \times A \rightarrow \mathbb{R}$, an initial state distribution $p(s_0)$ and transition dynamics $p(s_{t+1}|s_t,a_t)$. At each timestep $t$, the agent observes a state $s_t$, takes an action $a_t$, and receives a reward $r_t$. The agent chooses these actions using a policy $\pi$, which is a conditional distribution over actions given states. In this paper, we consider deterministic policies which map directly from states to actions $\pi: S \rightarrow A$. 

The discounted sum of future rewards is defined as the return $R_t = \sum_{i=t}^{T} \gamma^{(i-t)}r_i$ over some time horizon $T$ and with a discounting factor $\gamma \in [0,1]$. We define $\rho^\pi$ as the state visitation distribution when taking actions according to $\pi$. The goal in reinforcement learning is to learn a policy $\pi$ to maximize the expected return $J = \E_{s_i\sim \rho^\pi, a_i\sim\pi, r_i\sim E}{[R_0|s_0]}$.

The expected return when taking actions according to a specific policy $\pi$ is called the Q-function or action-value function, and is defined as:
\begin{equation}
    Q^\pi(s_t,a_t) = \E_{s_{i>t}\sim \rho^\pi, r_{i\ge t}\sim E}{[R_t|s_t,a_t]}
\end{equation}
which can be recursively stated as the Bellman equation:%(\cite{bellman1966dynamic})
\begin{equation}
    Q^\pi(s_t,a_t)=\E_{r_t,s_{t+1}\sim E}{[r_t + \gamma Q^\pi(s_{t+1},\pi(s_t,a_t))]}
\label{eq:bellman}
\end{equation}
Because $\pi$ is deterministic, the expectation in equation \ref{eq:bellman} depends only on the environment, allowing off-policy methods to learn $Q^\pi$ while using transitions generated with some other stochastic policy $\beta$.

\subsubsection{DDPG}
Our work uses the Deep Deterministic Policy Gradients algorithm (DDPG) (\cite{lillicrap2015continuous}), which is an off-policy, model-free reinforcement learning algorithm designed for use with deep neural networks in continuous action spaces. DDPG uses an actor-critic methodology. Two neural networks are trained: a critic $Q: S\times A \rightarrow \mathbb{R}$ parameterized by $\theta^Q$, 
%which learns to estimate the action-value function $Q^\pi$ using the Bellman equation, 
and an actor serving as the policy $\pi: S \rightarrow A$, which is updated using the policy gradient to directly maximize $Q^\pi$ with respect to the policy's parameters $\theta^\pi$:
\begin{equation}
            \nabla_{\theta^\pi} J = \E_{s_t\sim \rho^\beta}[\nabla_{\theta^\pi}Q(s_t,a|\theta^Q)|_{a=\pi(s_t|\theta^\pi)}]
\end{equation}
This quantity can be estimated with the following:
\begin{equation}
     \nabla_{\theta^\pi} J \approx \frac{1}{N}\sum_{i}\nabla_a Q(s_i,a|\theta^Q)|_{a=\pi(s_i) }\nabla_{\theta^\pi} \pi(s_i|\theta^\pi)
\end{equation}
The critic's parameters $\theta^Q$ are updated to minimize the loss:
\begin{equation}
    L^{crit} = \frac{1}{N}\sum_{\mkern0mu i}{(y_i-Q(s_i,a_i|\theta^Q))^2}
\end{equation}
where
\begin{equation}
    y_t = r_t + \gamma Q'(s_{t+1},\pi'(s_{t+1}))
\end{equation}
For stability, slower moving target networks $\pi'$ and $Q'$ are used to calculate $y_t$. These network's parameters are exponential moving averages of $\theta^\pi$ and $\theta^Q$ respectively.

DDPG maintains a replay buffer $R$ containing transition samples, which are tuples $(s_t, a_t, r_t, s_{t+1})$, and alternates between two stages. The first stage is to gather experience for $R$ by performing rollouts on the environment, choosing actions from a new policy $\beta = \pi + \epsilon$ where $\epsilon$ is random. 
%Additionally, every action usually has a small amount of noise added to it. 
The second stage is to train $\pi$ and $Q$ on batches of transition samples from $R$.

To efficiently gather experience, we run DDPG in parallel using multiple workers with synchronized copies of each network, averaging parameters across workers after each update. 

\begin{figure*}[!th]
  \centering
  \includegraphics[scale=0.25]{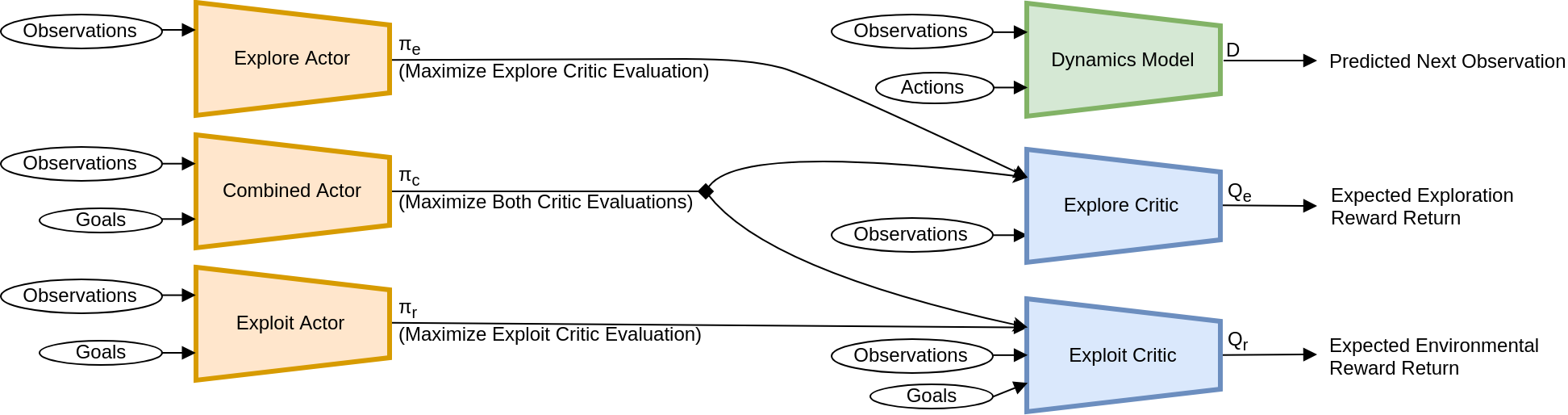}
  \caption{Forward Network Connections for DDPG+HER Learner with Curiosity-Driven Exploration. During testing, actions are taken by the exploit actor to maximize return on environmental rewards. During training, actions can be taken by any actor depending on which objectives we wish to emphasize.}
\end{figure*}
\subsubsection{DDPG with Goals}

In our work, we follow a goal based-framework. A goal $g \in G$ is sampled each episode, and $\pi$ and $Q$ are conditioned on these goals, making them $\pi: S \times G \rightarrow A$ and $Q: S \times A \times G \rightarrow \mathbb{R}$. Furthermore, the replay buffer instead stores transition samples as $(s_t||g, a_t, r_t, s_{t+1}||g)$, where the states are each concatenated with a goal. The environments' reward functions $r_t = r_{env}(s_{t+1},g)$ are also parameterized on whether a new state meets these goals.

\subsubsection{Hindsight Experience Replay}
In goal-based scenarios, hindsight experience replay increases the sample efficiency of replay buffer based algorithms like DDPG by adding additional augmented samples to the replay buffer. In doing so, HER allows the agent to evaluate its progress not only towards the goals that it was given by the environment, but also towards those that it actually reached in experience gathering rollouts, thus giving the agent hindsight.

HER acts by duplicating transition samples before placing them in the replay buffer, and in those duplicates, augmenting them by replacing the environment-provided goals with goals that were actually reached later in the same episode. HER requires the learning algorithm to have access to the reward function, and the rewards in the augmented samples are updated according to the newly replaced goals.

HER can also be implemented by expressly storing unmodified transition samples in the replay buffer and, with a certain probability, augmenting them when they are sampled from it. We use this method in our work.

\subsection{Environments}
%REWRITE/CUT
The block stacking environments that we consider in this work are based on the  Fetch robot environments from the OpenAI Gym API (\cite{plappert2018multi}) and are similar those used in (\cite{nair2018overcoming}). We test on separate environments for $n = 2$ to $4$ blocks. Each episode, target block locations are initialized in a stack somewhere on the surface of a table. The $n$ blocks are initialized at random locations on the table away from the target stack location. The blocks are uniquely labeled, and each block always goes to the same vertical position in the stack. The agent has $25n$ timesteps before the environment resets. These environments are fully observable, and observations include the claw state and full position, rotation, and velocity for both the robot's gripper and each block. These environments' goals specify the target positions of each block, and a block is considered correctly placed if its position is within an error tolerance $e$ from its target position. Actions are continuous and control the robot gripper’s movement in 3 dimensions as well as the state of the claw. 

Similar to (\cite{nair2018overcoming}), we consider two sparse reward formulations with these environments: binary and incremental. We can provide a single binary reward when the goal is fully achieved upon correctly placing all blocks:
\begin{equation}
    r_t^{binary} =  \begin{cases} 
            0 & \text{all blocks in place} \\
            -1 & \text{otherwise}
        \end{cases}
\end{equation}
We also consider incremental rewards for each block correctly placed:
\begin{equation}
    r_t^{incremental} = \text{no. of blocks in place} - \text{no. of blocks}
\end{equation}
%\begin{equation}
%    r_t^{incremental} = \sum_{i=1}^n\begin{cases} 
%            0 & \text{block $i$ correctly placed} \\
%            -1 & \text{otherwise}
%        \end{cases}
%\end{equation}

In both cases, we also add $1$ to the reward for moving the gripper away from the blocks once they are all correctly placed, but only the correct placement of all the blocks determines whether a goal has been achieved. 
%Both binary and incremental rewards are sparse, however the binary formulation requires far more initial exploration before a reward signal is likely to be encountered. 

\section{Methods}

To solve multi-block stacking with both incremental and binary rewards, we use three methods to improve the performance of a standard multi-worker DDPG+HER learner: curiosity-driven exploration, multi-criteria HER, and curriculum learning. 

First, we incorporate curiosity-driven exploration by training a forward dynamics model on state transitions visited by the agent and treating the dynamics model's prediction error on these transitions as an exploration reward. In order to have a certain portion of workers explore while others exploit, we train three separate policies to maximize exploration rewards, environmental rewards, and a weighted combination of both. Experience from rollouts is shared among each network regardless of which policy collected it. By doing so, we can use different policies at training time than at test time.

% To enable high flexibility as to which objectives we pursue during experience gathering and testing, we train two critics and three actors. One critic is trained to predict expected return from environment reward, and the other is trained to predict expected return from exploration reward. Likewise, two of the actors are trained to choose actions which maximize each of these action-value functions, and the third is trained to maximize the weighted average of both.

% As a hyperparameter, we can also optionally assign different policies to individual workers such that an effective and distinct mix of exploration and exploitation objectives are pursued when collecting experience.

Second, we introduce a form of hindsight experience replay better suited for multi-criteria goal-based environments, where a criteria in our environment is defined as the position of a specific block. Our method randomly performs the goal replacement operation on each independent criteria in a goal rather than on an entire goal at once, decoupling the individual effects of each criteria on the reward function and providing higher sample efficiency. 

Third, we use curriculum learning by training the agent on two easier skill-building environments before training on the target multi-block stacking task.

\subsection{Curiosity Driven Exploration with Multiple Policies}

%Because reward is sparse and difficult to discover in multi-block stacking, 
We use curiosity-driven exploration to encourage an agent to visit transitions which are novel and surprising to it. We define an auxiliary exploration reward in addition to environmental reward, and we train separate critics for each. The explore critic $Q_e$  predicts the action-value function for exploration rewards, and the exploit critic $Q_r$ predicts the action-value function for environmental rewards. We train three actor policies $\pi_e, \pi_r, \pi_c$ which respectively maximize exploration rewards, environmental rewards, and a weighted combination of both. By training separate polices, we can make our agent pursue multiple and various objectives at training time and maximize only environmental rewards at test time.

We maintain a forward dynamics neural network $D:S\times A \rightarrow S$ parameterized by $\theta^d$ to predict the next observation given the current observation and action, and we train it on the same transition samples from the replay buffer as our agent at each DDPG update step. For each transition sample trained on, an exploration reward for the sample is defined as the squared error between the predicted next state $D(s_t,a_t|\theta^{d})$ and the actual next state $s_{t+1}$. The minibatch loss function for $D$ and the exploration reward is formulated as:
\begin{equation} 
L_{d} = \frac{1}{N}\sum_{\mkern0mu i}r_i^{explore} = \frac{1}{N}\sum_{\mkern0mu i}(s_{i+1}-D(s_i,a_i|\theta^{d}))^2
\end{equation}
By passing this error $r_i^{explore}$ to our agent as an exploration reward to maximize, we encourage our agent to pursue transitions that are difficult to predict and unlike transitions currently in the replay buffer. 

% We refer to the actor-critic pair that is trained on this reward as the explore actor-critic pair. 
%Ideally, along the way, it will discover new sources of environmental reward.

% To allow our DDPG agent to pursue this exploration reward when we could like, and ignore it at other times, two actor-critic pairs are trained on the same data. To accomplish exploitation purposes, $Q_r:S\times G\times A \rightarrow \mathbb{R}$ and $\pi_{r}:S\times G \rightarrow A$ are trained to respectively predict and maximize expected return from environmental rewards. On the same note, to accomplish exploration purposes, $Q_e:S\times A \rightarrow \mathbb{R}$ and $\pi_{e}:S \rightarrow A$ are trained to predict and maximize the expected return from exploration rewards. Additionally, a third actor $\pi_{c}:S\times G \rightarrow A$ is trained on the same data to maximize the combination of both action-value functions with equal importance.

With two separate reward sources, we group our multiple actors and critics into two DDPG actor-critic pairs. On exploration reward, we train $Q_e$ and $\pi_e$ as our explore actor-critic pair. On environmental reward, we train $Q_r$ and $\pi_r$ as our exploit actor-critic pair. In addition to these actor-critic pairs, we also train $\pi_c$ as our combined actor. $\pi_c$ pursues both exploration and environmental reward by maximizing a weighted average of both critics' action-value functions.

We train multiple actors towards different objectives so that we can assign a portion of our workers to follow an exploration related policy $\pi_e$ or $\pi_c$ while the rest follow the exploit policy $\pi_r$. Doing so allows us to diversify the experience gathered and make less sacrifices toward either exploration or exploitation objectives than if we were to only ever choose actions which maximize a weighted combination of the two. With multiple actors, we can specialize our workers and maximize both environmental and exploration rewards when gathering experience and then use $\pi_{r}$ at test time to solely maximize environmental rewards. Below we describe these networks in more detail.

\subsubsection{Exploit Actor and Critic}

$\pi_{r}:S\times G \rightarrow A$ and $Q_r:S\times G\times A \rightarrow \mathbb{R}$ together form the exploit actor-critic pair, which is trained on the normal DDPG+HER goal-based RL objective for maximizing return on environmental rewards conditioned on goals. This actor-critic pair follows the same configuration and update rules as what would be used in vanilla DDPG+HER. The loss function for $Q_{r}$ to minimize with respect to its parameters $\theta^Q_r$ is:
\begin{equation}
    L_{r}^{crit} =\E_{s_t\sim\rho^\beta,a_t\sim\beta,r_t\sim E}(y_t^{r}-Q_{r}(s_t,a_t,g|\theta^Q_r))^2
\end{equation}

where $y_t^{r}$ is calculated using the target exploit actor and critic $\pi'_r$ and $Q'_{r}$:
\begin{equation}
    y_t^{r} = r_t^{env} + \gamma Q'_{r}(s_{t+1},\pi'_r(s_{t+1}, g),g)
\end{equation}

$\pi_r$ is updated using the standard goal-based DDPG policy gradient to maximize $Q_r$ with respect to $\pi_r$'s parameters $\theta^\pi_r$.
% \begin{equation}
%             %\nabla_{\theta^\pi_{r}} J \approx \frac{1}{N}\sum_{i}
%             \nabla_{\theta^\pi_{r}}J = \E_{s_t\sim\rho^\beta}[\nabla_{\theta^\pi_{r}}Q_{r}(s_t,a,g|\theta^Q_r)|_{a=\pi_r(s_t, g|\theta^\pi_r)}]
% \end{equation}

\subsubsection{Explore Actor and Critic}

$\pi_{e}:S \rightarrow A$ and $Q_e:S\times A \rightarrow \mathbb{R}$ together form the explore actor-critic pair, which is trained on the objective of maximizing return on exploration reward. Goals do not affect exploration rewards and are not factored in these calculations. The loss function for $Q_{e}$ to minimize with respect to its parameters $\theta^Q_e$ is:
\begin{equation}
    L_{e}^{crit} =\E_{s_t\sim\rho^\beta,a_t\sim\beta,r_t\sim E}(y_t^{e}-Q_{e}(s_t,a_t|\theta^Q_e))^2
\end{equation}
where $y_t^{e}$ is calculated using the target explore actor and critic $\pi'_e$ and $Q'_{e}$:
\begin{equation}
    y_t^{e} = r_t^{explore} + \gamma Q'_{e}(s_{t+1},\pi'_e(s_{t+1}))
\end{equation}
Likewise, $\pi_e$ is updated using the standard DDPG policy gradient to maximize $Q_e$ with respect to $\pi_e$'s parameters $\theta^\pi_e$.
%\begin{equation}
%            \nabla_{\theta^\pi_{e}}J = %\E_{s_t\sim\rho^\beta}[\nabla_{\theta^\pi_{e}}Q_{e}(s_t,a|\theta^Q_e)|_{a=\pi_e(s_t|\theta^\pi_e)}]
%\end{equation}

\subsubsection{Combined Actor and POP-ART}
Once our agent has an idea of how to find environmental rewards, it is usually more advantageous to explore trajectories close to what actually results in those rewards.
Towards this end, we train our combined actor $\pi_{c}:S\times G \rightarrow A$ to choose actions that maximize both the exploration and exploitation objectives simultaneously. $\pi_{c}$ outputs actions that maximize the weighted combination of both $Q_e$ and $Q_r$'s action-value functions. 

We intend to maintain a normalized scale at which to compare the return estimates from $Q_e$ and $Q_r$ so that we can intuitively weight their relative importance to $\pi_c$. We also need to account for the fact that the magnitude of both action-value functions may change drastically over the course of training. This is especially true of $Q_{e}$ which predicts the return from the moving exploration reward function. To accomplish both of these goals, each of the targets $y^{e}$ and $y^{r}$ for $Q_e$ and $Q_r$ are adaptively normalized such that we also maintain normalized versions $n_e^Q$ and $n_r^Q$ of both action-value functions with the same relative scale at all times. We can then intuitively weight the relative importance of $n_e^Q$ and $n_r^Q$ for $\pi_c$ to maximize. In our case, we weight them equally.

To do this, we use PopArt normalization (\cite{van2016learning}), which allows us to adaptively normalize our critics' targets without hurting the accuracy of our predictions. Here we only sketch PopArt informally. See (\cite{van2016learning}) for more details. For each critic target $y^e$ and $y^r$ we keep an online estimate of its mean and standard deviation $\sigma_e, \mu_e$ and $\sigma_r, \mu_r$. We then parameterize $Q_e$ and $Q_r$ as linear transformations of the suitably normalized action-value functions $n_e^Q$ and $n_r^Q$:
\begin{equation}
\label{eq:popart}
\begin{split}
      Q_e(s,a|\theta^Q_e) &= \sigma_e n_e^Q(s,a|\theta^Q_e) + \mu_e\\
      Q_r(s,a,g|\theta^Q_r) &= \sigma_r n_r^Q(s,a,g|\theta^Q_r) + \mu_r 
\end{split}
\end{equation}
$n_e^Q$ and $n_r^Q$ are the actual networks that we train parameterized by $\theta^Q_e$ and $\theta^Q_r$, and when the statistics $\sigma_e, \mu_e$ and $\sigma_r, \mu_r$ are updated, the top layers of $n_e^Q$ and $n_r^Q$ are also adjusted to preserve equation \ref{eq:popart}. Similarly, our target critics $Q_e', Q_r'$ are equivalent linear transformations of target normalized critic networks ${n_e^Q}', {n_r^Q}'$.
% \begin{equation}
% \label{eq:popart}
% \begin{split}
%       Q_r'(s,a,g) &= \sigma_r {n_r^Q}'(s,a,g) + \mu_r \\
%       Q_e'(s,a) &= \sigma_e {n_e^Q}'(s,a) + \mu_e
% \end{split}
% \end{equation}

In our implementation, the scale-invariant loss functions for each of our two critic networks $n_e^Q$ and $n_r^Q$ are:
\begin{equation}
    L_{e}^{crit} = \frac{1}{N}\sum_{\mkern0mu i}{(\frac{y_i^{e}-\mu_j}{\sigma_e}-n_e^Q(s,a|\theta^Q_e))^2}
\end{equation}
\begin{equation}
    L_{r}^{crit} = \frac{1}{N}\sum_{\mkern0mu i}{(\frac{y_i^{r}-\mu_j}{\sigma_r}-n_r^Q(s,a,g|\theta^Q_r))^2}
\end{equation}

\begin{algorithm}[h]
\SetAlgoLined
\DontPrintSemicolon
\kwGiven{Worker policies $\pi_0,\pi_1,...,\pi_w | \pi_i \in \{\pi_e,\pi_r,\pi_c\}$}
 Randomly initialize networks $D, n^Q_e, n^Q_r, \pi_e, \pi_r, \pi_c$\;
 Initialize target networks ${n_e^Q}', {n_r^Q}', \pi_e', \pi_r', \pi_c'$\;
 \;
 \emph{(Execute for each parallel worker $i$):}\;
 Initialize replay buffer $R$\;
 \For{Epoch = $1, ..., E$}{
    \For{Cycle = $1, C$}{
     \For{Episode = $1, M$}{
        Sample $\epsilon$ and set $\beta \gets \pi_i + \epsilon$\;
        Receive initial state $s_0$ and goal $g$\;
        \For{t = $0, T$}{
            Select action $a_t = \beta(s_t,g)$ with noise\;
            Take action $a_t$, receive $r_t, s_{t+1}$\;
            Store $(s_t||g, a_t, r_t, s_{t+1}||g)$ in $R$\;
        }
    }
    \For {Batch = $1, ..., K$}{
        Sample batch $B$ from $R$ with HER augmentations\;
        Train $D$ on $B$\;
        \ForEach{\upshape transition sample $j$ in $B$}{
            Set $r_j^{explore}$ and add it to sample\;
        }
        Train $n^Q_e, n^Q_r, \pi_e, \pi_r, \pi_c$ on $B$\;
        Update target networks\;
        Average network parameters over workers\;
    }
    }
    Test performance on episodes using $\pi_r$\;
 }

 \caption{DDPG+HER with Curiosity}
\end{algorithm}

By training $n_e^Q$ and $n_r^Q$ to predict normalized action-value functions, we can update $\pi_c$ to jointly maximize the evaluation from both the explore and exploit critics with equal importance:
\begin{equation}
            \nabla_{\theta^\pi_{c}}J = \E_{s_t\sim\rho^\beta}[\nabla_{\theta^\pi_{c}}Q_{c}^{norm}(s_t,a,g)|_{a=\pi_c(s_t, g|\theta^\pi_c)}]
            % \nabla_{\theta^\pi_{c}}J \approx \E_{s_t\sim E}[\nabla_{\theta^\pi_{c}}Q_c^{norm}(s_t,a,g)|_{a=\pi'_c(s_t, g)}]
\end{equation}
where
\begin{equation*}
    Q_c^{norm}(s_t,a,g) = \frac{n_e^Q(s_t,a|\theta^Q_{e}) + n_r^Q(s_t,a,g|\theta^Q_{r})}{2}
\end{equation*}

Then for our three actors $\pi_c, \pi_e, \pi_r$ the implemented policy gradient update rules are:
\begin{equation}
            \nabla_{\theta^\pi_{c}}J \approx \frac{1}{N}\sum_{i}\nabla_aQ_{c}^{norm}(s_i,a,g)\nabla_{\theta^\pi_{c}}a|_{a=\pi_{c}(s_i,g|\theta^\pi_c)}
\end{equation}
\begin{equation}
            \nabla_{\theta^\pi_{e}}J \approx \frac{1}{N}\sum_{i}\nabla_a n_e^Q(s_i,a|\theta^Q_e)\nabla_{\theta^\pi_{e}}a|_{a=\pi_{e}(s_i|\theta^\pi_e)}
\end{equation}
\begin{equation}
            \nabla_{\theta^\pi_{r}}J \approx \frac{1}{N}\sum_{i}\nabla_a n_r^Q(s_i,a,g|\theta^Q_r)\nabla_{\theta^\pi_{r}}a|_{a=\pi_{r}(s_i,g|\theta^\pi_r)}
\end{equation}

In our experiments, when we used curiosity-driven learning, we chose actions using the combined policy $\pi_c$ instead of the pure explore policy $\pi_e$. The pure explore policy $\pi_e$ is still useful to train the explore critic $Q_e$ which is then used to train the combined policy $\pi_c$.  
%We found that $\pi_c$ performed no worse than $\pi_e$ in the absence of environmental rewards, with the additional advantage of exploring trajectories that are also likely to yield these environmental rewards once the agent has begun to find them. 
%$\pi_e$ is still trained so that we may calculate $Q_e$ with respect to the optimal policy for gathering exploration rewards.

\subsection{Multi-Criteria Hindsight}
We define the multiple criteria in a goal as the individual target block positions that the goal specifies. In general, for other environments, criteria can be elements of a goal that require learning separate skills to accomplish. To increase the quality of data provided by hindsight experience replay, we randomly perform the hindsight goal replacement operation independently on each criteria in a goal that we are augmenting. This is done instead of replacing the entire goal with one reached later in the same episode.

Our method provides more transition samples to the agent with goals that are only partially completed later in the same episode. With normal HER, all hindsight augmented samples that the agent receives contain goals in which all criteria were satisfied at a later timestep. With multi-criteria HER, the agent will still receive a portion of goals that it later satisfied completely, and it will also receive many goals that it later only satisfied some criteria for. 

% By providing these partially completed goals, we speculate that the agent receives higher quality information, as experiments show that multi-criteria hindsight provides a boost to inter-task generalization.

\begin{algorithm}[h]
\SetAlgoLined
\DontPrintSemicolon
\kwGiven{
    \begin{itemize}\setlength{\itemsep}{-3pt}
    %   \item a transition sample $(s_t||g, a_t, r_t, s_{t+1}||g)$
       \item an augmentation probability $z$
       \item a Replay Buffer $R$
    \end{itemize}
}
Sample a batch $B$ from $R$\;
\ForEach{\upshape transition sample $(s_t||g, a_t, r_t, s_{t+1}||g)$ in $B$}{
    \ForEach{\upshape target block position $p_i$ in $g$}{
        Sample $u \sim U(0,1)$\;
        \uIf{$u < z$}{
            Sample a position $p'_i$ that block $i$ reached later in the same episode.\;
        }
        \Else{
            $p_i' \gets p_i$\;
        }
    }
    $g' \gets p_0'||p_1'||...||p_n'$\;
    $r_t' \gets r(s_{t+1},g')$\;
    replace transition sample w/ $(s_t||g', a_t, r_t', s_{t+1}||g')$
}
Pass $B$ with augmented transition samples to neural networks for training
 \caption{Multi-Criteria HER Augmentation Step}
\end{algorithm}

In our experiments, for both binary and incremental reward formulations, using multi-criteria HER results in significant, if not critical, improvements to sample efficiency and inter-task generalization.

\subsection{Curriculum}
Although multi-criteria hindsight sampling allows for more sample-efficient learning and curiosity driven exploration assists in reward discovery, it was necessary to employ curriculum learning to successfully solve multi-block stacking with  sparse rewards. Training was broken into three stages, in which reaching a threshold success rate in a previous stage caused the agent to transition to the next stage. At the beginning of each stage, the DDPG algorithm was restarted, transferring only the weights of each network from a previous stage and reinitializing an empty replay buffer.

In \emph{stage 1}, the agent trains on a non-stacking version of the block environment to help it learn fundamental skills that are transferable to the target block stacking task. The stage 1 environment is initialized with the same number of randomly placed blocks as the target stacking task. Each episode, rather than in a stack, the blocks' target positions are randomly placed on the surface of the table. A single block's target position may also be in the air instead. This stage is designed to provide less challenging tasks in which the agent can more easily discover the basic block manipulation mechanics necessary for completing the harder stacking task.  

In \emph{stage 2}, the agent trains on actual block stacking with the environment initialized at various intermediate stages of completion. At each episode, a random number of the n blocks between $0$ and $n-1$ are initialized already in the correct position on the stack. Some targets may also still be on the table rather than on the stack.

Finally, in \emph{stage 3}, the agent trains on the target block stacking task, in which all blocks were consistently initialized on the table, away from their target locations on the stack.

\begin{figure}[t]
  \centering
  \includegraphics[scale=0.35]{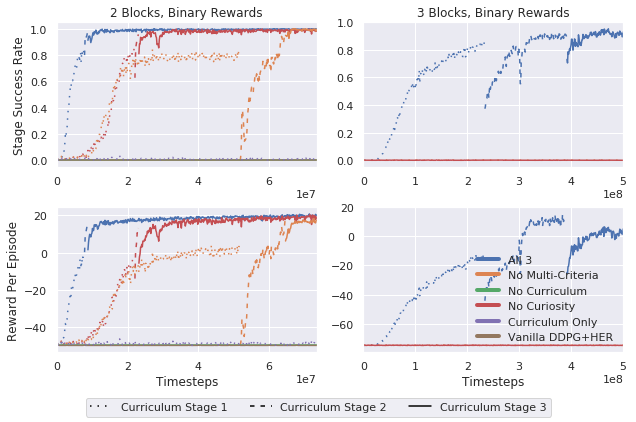}
  \caption{Success rates and per-episode rewards for block stacking with binary rewards. Success rates and per-episode reward values shown here are for the respective curriculum stage's task in which they are measured. }
  \label{fig:bin2_4}
\end{figure}

\section{Experiments}

\begin{figure*}[ht]
  \centering
  \includegraphics[scale=0.35]{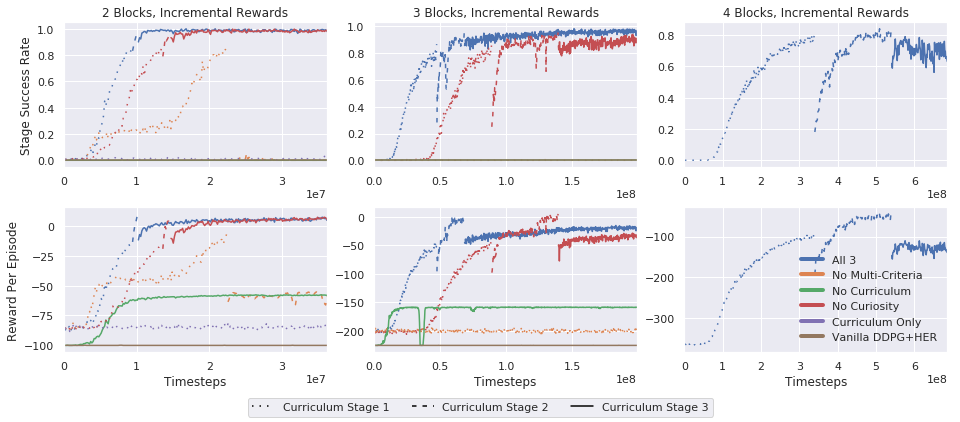}
  \caption{Success rates and per-episode rewards for block stacking with incremental rewards.  Success rates and per-episode reward values shown here are for the respective curriculum stage's task in which they are measured.}
  \label{fig:inc2_4}
\end{figure*}

In this section, we show our method's performance on the block stacking tasks using both binary and incremental rewards. Stacking 2, 3, and 4 blocks were tested. Ablations are also shown to demonstrate the effectiveness of each our methods. We performed tests using the following configurations:

\emph{All 3}: Multi-criteria HER, curiosity-driven exploration, and curriculum learning are all used with our DDPG+HER learner.

\emph{No Curiosity}: Multi-criteria HER and curriculum learning are used, but all actions are chosen using $\pi_r$. An explore actor-critic and combined actor are not trained.

\emph{No Multi-Criteria}: Curiosity-driven exploration and curriculum learning are used, however HER is done in the original way as defined in (\cite{andrychowicz2017hindsight}).

\emph{No Curriculum}:  Multi-criteria HER and curiosity-driven exploration are used, however the agent only trains on stage 3 of the curriculum, which is the actual target task of multi-block stacking.

\emph{Vanilla DDPG+HER}: None of the three methods introduced in section 2 are used. This is the original DDPG+HER algorithm as in (\cite{andrychowicz2017hindsight}). All actions are chosen using $\pi_r$.

We trained our agent using 8 to 32 parallel workers depending on the difficulty of the task. When curiosity driven-exploration was used, during experience gathering rollouts, we assigned half of the workers to take actions using $\pi_c$, and the other half using $\pi_r$. Also during experience gathering, we applied parameter-space noise (\cite{plappert2017parameter}) to the actor networks used and gaussian noise to the actions chosen. Comprehensive hyper-parameter details can be found in the supplementary materials associated with this paper.

Success rates and per-episode reward were measured during discrete testing phases in every epoch of training. During testing, actions were always chosen using $\pi_r$. An episode was considered successful if its goal $g$ was achieved during the episode's final state $s_T$.

Success rate and per-episode reward statistics were a moving average over the last 100 episodes tested on. These two statistics are shown as a function of total environment interaction timesteps for binary reward tasks in Figure \ref{fig:bin2_4} and for incremental reward tasks in Figure \ref{fig:inc2_4}.

\begin{table}[H]
\captionsetup{justification=centering}
\caption{Highest Success Rates with Binary Rewards over 100 Episode Sliding Window}
\label{bin_table}
\begin{center}
\begin{tabular}{llll}
\multicolumn{1}{c}{\bf Method}  &\multicolumn{1}{c}{\bf Stack-2} &\multicolumn{1}{c}{\bf Stack-3} &\multicolumn{1}{c}{\bf Stack-4}\\
\hline \\
All-3  & \textbf{1.00} & \textbf{0.95} & 0.00\\
No Curiosity  & \textbf{1.00} & 0.00 & -\\
No Multi-Criteria  & \textbf{1.00} & 0.00 & -\\
No Curriculum  & 0.00 & - & -\\
Curriculum  Only & 0.00 & - & -\\
Vanilla DDPG+HER  & 0.00 & - & - \\
\end{tabular}
\end{center}
\end{table}

\begin{table}[H]
\captionsetup{justification=centering}
\caption{Highest Success Rates with Incremental Rewards over 100 Episode Window}\label{inc_table}
\begin{center}
\begin{tabular}{llll}
\multicolumn{1}{c}{\bf Method}  &\multicolumn{1}{c}{\bf Stack-2} &\multicolumn{1}{c}{\bf Stack-3} &\multicolumn{1}{c}{\bf Stack-4}\\
\hline \\
All-3  & \textbf{1.00} & \textbf{0.98} & \textbf{0.79} \\
No Curiosity  & \textbf{0.99} & \textbf{0.94} & -\\
No Multi-Criteria  & 0.00 & 0.00 & -\\
No Curriculum  & 0.00 & 0.00 & - \\
Curriculum Only & 0.00 & - & -\\
Vanilla DDPG+HER  & 0.00 & - & - \\
\end{tabular}
\end{center}
\end{table}

Tables \ref{inc_table} and \ref{bin_table} show the highest success rates for each method on the target bock stacking tasks with binary and incremental reward formulations. For methods that used a curriculum but did not reach the target task in the third stage, the final network weights were used to test performance at the target block stacking task anyways.

% Plots of success rates and test-phase reward-per-episode for incremental reward block stacking can also be seen in Figure ~\ref{fig:inc2_5}. 

Vanilla DDPG+HER was unable to solve block-stacking with any number of blocks and either reward formulation. 

Stacking 2 blocks with either reward formulation was solvable as long as the agent trained on the curriculum and used multi-criteria HER. Using curiosity-driven exploration without multi-criteria HER allowed the agent to make progress on stage 1 of the curriculum, but when incremental rewards were given, it failed to generalize between the stage 1 task and the stage 2 task well enough to continue learning.

Stacking 3 blocks with incremental rewards required the use of both curriculum learning and multi-criteria HER to solve. With binary rewards, stacking 3 blocks required the use of all three methods, as curiosity-driven exploration was necessary to find a reward signal. 

Due to limits on computational resources, stacking 4 blocks was only tested with all three methods to measure the best possible performance. No progress was made on the binary reward environment, and in the incremental reward environment, a max success rate of $0.79$ was reached on the target block stacking task.

% \subsubsection*{Multi-Criteria HER Curiosity-Driven Exploration}
Multi-criteria HER provided clear improvements to sample efficiency, and was necessary for stacking three or more blocks.

% \subsubsection*{Curiosity-Driven Exploration}
Agents with curiosity-driven exploration learned to solve tasks with less environment interactions than those without. With incremental rewards, block stacking was easy enough to be solved without curiosity-driven exploration, however with binary rewards, curiosity was required to solve stacking 3 blocks. 

Finally, curriculum learning was necessary for any of the stacking tasks, as no method could progress on the target stacking task without first training on stages 1 and 2.

\section{Conclusion}

%In this paper, we propose the use of three methods to tackle sparse reward block stacking tasks previously only solvable with help from demonstration. 
By combining curiosity-based exploration with curriculum learning and multi-criteria HER, we are the first to solve sparse reward multi-block stacking without demonstrations. This work shows that even very challenging sparse reward environments can be solved through a combination of existing techniques. In future work, other methods of intrinsic exploration such as Go-Explore (\cite{go-explore}) might prove more effective than curiosity-driven exploration when combined with HER. In our work, we generate curricula in a hand-designed way based on domain knowledge. This might not be possible in more complex domains such as real-world robotics. Because of this, further research in automatically generating curricula is likely to be a fruitful direction when combined with HER.

\subsubsection*{References}

\printbibliography[heading=none]

\onecolumn
\appendix

\section{Links}
A video showcasing this project is available at \\ \url{https://youtu.be/stZX4o0H8Ro} 

Code for our modified DDPG Learner is available at \\ \url{https://github.com/CDMCH/ddpg-with-curiosity-and-multi-criteria-her}\\ 
and code for our block stacking environments is available at\\ \url{https://github.com/CDMCH/gym-fetch-stack}

Our DDPG learner uses code modified from the OpenAI baselines repository (\cite{dhariwalopenai}).

\section{Experiment Details}

Observation and goal network inputs were normalized to have a mean of zero and standard deviation of one. Once normalized, they were also clipped to the range [-5, 5].

All networks were fully connected with 3 hidden layers and 256 hidden units in each layer. Hidden layers used ReLU activations, while the output layers of actor networks used tanh. The action space was re-scaled to the fit the tanh range of [-1, 1], and to prevent vanishing gradients, the preactivations of the actor output layers were penalized by the square of their magnitude with a coefficient of 0.001.

The DDPG algorithm was run in parallel using multiple message passing interface (MPI)-based workers. Network parameters and normalization statistics were averaged across workers during update steps. The actor policy, $\pi_e$, $\pi_r$, or $\pi_c$ that each worker used during experience gathering was set as a hyperparameter. All workers used $\pi_r$ during performance testing. Different worker amounts were used depending on the difficulty of the task:

\begin{table}[h]
\captionsetup{justification=centering}
\caption{Parallel Worker Amounts by Task}
\label{worker_table}
\begin{center}
\begin{tabular}{llll}
\multicolumn{1}{c}{\bf Task}  &\multicolumn{1}{c}{\bf Number of MPI Workers} \\
\hline \\
Stack 2, Sparse Rewards & 8\\
Stack 3, Sparse Rewards & 32 \\
Stack 4, Sparse Rewards & 32 \\
Stack 2, Incremental Rewards & 8 \\
Stack 3, Incremental Rewards & 8\\
Stack 4, Incremental Rewards & 32 \\
\end{tabular}
\end{center}
\end{table}

\newpage
The following hyperparameters were used in our experiments:

\begin{table}[H]
\captionsetup{justification=centering}
\caption{Hyperparameters for Block Stacking Tasks}
\label{hyp_table}
\begin{center}
\begin{tabular}{ll}
\multicolumn{1}{c}{\bf Hyperparameter}  &\multicolumn{1}{c}{\bf Value} \\
\hline \\
Optimizer & Adam (\cite{kingma2014adam})\\
$n_r$ Learning Rate & 0.001 \\
$n_r$ L2 Regularization Coefficient & 0 \\
$\pi_r$ Learning Rate & 0.001 \\
Target Exploit Actor-Critic Polyak-averaging Coefficient & 0.001 \\
$n_e$ Learning Rate & 0.001 \\
$n_e$ L2 Regularization Coefficient & 0.01 \\
$\pi_e$ Learning Rate & 0.001 \\
Target Explore Actor-Critic Polyak-averaging Coefficient: & 0.05 \\
$\pi_c$ Learning Rate & 0.001 \\
$\pi_c$ Explore vs Exploit Critic Weighting & $0.5, 0.5$ \\
$D$ Learning Rate & 0.007 \\
Episode Time Horizon & $50 * \text{num blocks}$ \\
$\gamma$ & $1 - 1/{\text{episode time horizon}}$ \\
MPI Worker Replay Buffer Size & $10^6$ transitions \\
Parameter Space Noise $\sigma$ Target & 0.1 \\
Guassian Action Noise $\sigma$ & 0.04 \\
Traditional HER Augmentation Probability (when used) & 0.8\\
Multi-Criteria HER Augmentation Probability (when used) & 0.8 \\
Cycles per Epoch & 50\\
Experience Gathering Episodes per Cycle & 8 (per MPI worker) \\
Training Batches per cycle & 8 \\
Network Update Batch Size & 1024 transitions (per MPI worker)\\
Test Episode Rollouts Per Epoch & 50\\
\end{tabular}
\end{center}
\end{table}

\end{document}